\documentclass[american]{iopjournal}
\usepackage[T1]{fontenc}
\usepackage[utf8]{inputenc}
\usepackage{array}
\usepackage{refstyle}
\usepackage{float}
\usepackage{units}
\usepackage{url}
\usepackage{algorithm2e}
\usepackage{amsmath}
\usepackage{graphicx}

\makeatletter

\AtBeginDocument{\providecommand\figref[1]{\ref{fig:#1}}}
\AtBeginDocument{\providecommand\subsecref[1]{\ref{subsec:#1}}}
\AtBeginDocument{\providecommand\Figref[1]{\ref{Fig:#1}}}
\AtBeginDocument{\providecommand\tabref[1]{\ref{tab:#1}}}
\providecommand{\tabularnewline}{\\}
\RS@ifundefined{subsecref}
  {\newref{subsec}{name = \RSsectxt}}
  {}
\RS@ifundefined{thmref}
  {\def\RSthmtxt{theorem~}\newref{thm}{name = \RSthmtxt}}
  {}
\RS@ifundefined{lemref}
  {\def\RSlemtxt{lemma~}\newref{lem}{name = \RSlemtxt}}
  {}

\usepackage{xcolor}
\definecolor{mymaroon}{cmyk}{0, 0.87, 0.68, 0.32}
\definecolor{myhalfgray}{gray}{0.55}
\definecolor{myframe}{RGB}{207, 207, 207}
\definecolor{mybg}{RGB}{250, 250, 250}
\definecolor{myred}{RGB}{186, 33, 33}
\definecolor{mygreen}{RGB}{0, 128, 0}
\definecolor{mycyan}{RGB}{64, 128, 128}
\definecolor{mypurple}{RGB}{170, 34, 255}
\definecolor{myorange}{RGB}{206, 92, 0}
\definecolor{myblue}{RGB}{32, 74, 135}

\usepackage{listings}
\lstdefinelanguage{myPython}{
    morekeywords={access,and,as,break,class,continue,def,del,elif,
                  else,except,exec,finally,for,from,global,if,import,
                  in,is,lambda,not,or,pass,print,raise,return,try,while},
    morekeywords=[2]{abs,all,any,basestring,bin,bool,bytearray,callable,
                     chr,classmethod,cmp,compile,complex,delattr,dict,dir,
                     divmod,enumerate,eval,execfile,file,filter,float,format,
                     frozenset,getattr,globals,hasattr,hash,help,hex,id,input,
                     int,isinstance,issubclass,iter,len,list,locals,long,map,
                     max,memoryview,min,next,object,oct,open,ord,pow,property,
                     range,raw_input,reduce,reload,repr,reversed,round,set,
                     setattr,slice,sorted,staticmethod,str,sum,super,tuple,type,
                     unichr,unicode,vars,xrange,zip,apply,buffer,coerce,intern},
	keywordstyle=\color{mygreen}\ttfamily\textbf,
	keywordstyle=[2]\color{myblue}\ttfamily\textbf,
    sensitive=true,
    morecomment=[l]\#,
    morestring=[b]',
    morestring=[b]",
    morestring=[s]{'''}{'''},
    morestring=[s]{"""}{"""},
    morestring=[s]{r'}{'},
    morestring=[s]{r"}{"},
    morestring=[s]{r'''}{'''},
    morestring=[s]{r"""}{"""},
    morestring=[s]{u'}{'},
    morestring=[s]{u"}{"},
    morestring=[s]{u'''}{'''},
    morestring=[s]{u"""}{"""},
    literate=
    {+=}{{{+=}}}1
    {-=}{{{-=}}}1
    {*=}{{{$^\ast$=}}}1
    {/=}{{{/=}}}1,
    literate=
    *{-}{{{\color{myorange}-}}}1
	{+}{{{\color{myorange}+}}}1
    {*}{{{\color{myorange}$\ast$}}}1
    {/}{{{\color{myorange}/}}}1
	{=}{{{\color{myorange}=}}}1
    {^}{{{\color{myorange}\^{}}}}1
    {?}{{{\color{myorange}?}}}1,
    identifierstyle=\color{black}\ttfamily,
    commentstyle=\color{mycyan}\ttfamily\textit,
    stringstyle=\color{myred}\ttfamily,
    showspaces=false,
    showstringspaces=false,
    rulecolor=\color{myframe},
    frame=single,
    frameround={t}{t}{t}{t},
    framexleftmargin=6mm,
    numbers=left,
    numberstyle=\tiny\color{myhalfgray},
    backgroundcolor=\color{mybg},
}

\SetAlgorithmName{Listing}{}{}

\usepackage{tabularx} 
\usepackage{multirow}
\usepackage{colortbl}
\usepackage{dsfont}

\makeatother

\usepackage{babel}
\usepackage[backend=bibtex8,sorting=none,url=false,isbn=false,date=year]{biblatex}
\begin{document}
 \renewbibmacro{in:}{}

\articletype{Paper} 

\title{A flexible framework for structural plasticity in GPU-accelerated sparse spiking neural networks}

\author{James C. Knight$^1$\orcid{0000-0003-0577-0074}, Johanna Senk$^{1,2,*}$\orcid{0000-0002-6304-062X} and Thomas Nowotny$^{1}$\orcid{0000-0002-4451-915X}}

\affil{$^1$Sussex AI, School of Engineering and Informatics, University of Sussex, Brighton, United Kingdom}

\affil{$^2$Institute for Advanced Simulation (IAS-6), Jülich Research Centre, Jülich, Germany}

\affil{$^*$Author to whom any correspondence should be addressed.}

\email{j.senk@sussex.ac.uk}

\begin{abstract}
The majority of research in both training Artificial Neural Networks
(ANNs) and modeling learning in biological brains focuses on synaptic
plasticity, where learning equates to changing the strength of existing
connections. However, in biological brains, structural plasticity
-- where new connections are created and others removed -- is also
vital, not only for effective learning but also for recovery from
damage and optimal resource usage. Inspired by structural plasticity,
pruning is often used in machine learning to remove weak connections
from trained models to reduce the computational requirements of inference.
However, the machine learning frameworks typically used for backpropagation-based
training of both ANNs and Spiking Neural Networks (SNNs) are optimized
for dense connectivity, meaning that pruning does not help reduce
the training costs of ever-larger models. The GeNN simulator already
supports efficient GPU-accelerated simulation of sparse SNNs for computational
neuroscience and machine learning. Here, we present a new flexible
framework for implementing GPU-accelerated structural plasticity rules
and demonstrate this first using the e-prop supervised learning rule
and DEEP R to train efficient, sparse SNN classifiers and then, in
an unsupervised learning context, to learn topographic maps. Compared
to baseline dense models, our sparse classifiers reduce training time
by up to $10\times$ while the DEEP R rewiring enables them to perform
as well as the original models. We demonstrate topographic map formation
in faster-than-realtime simulations, provide insights into the connectivity
evolution, and measure simulation speed versus network size. The proposed
framework will enable further research into achieving and maintaining
sparsity in network structure and neural communication, as well as
exploring the computational benefits of sparsity in a range of neuromorphic
applications.
\end{abstract}

\section{Introduction}

\label{sec:introduction}

A simulation of a Spiking Neural Network~(SNN) is usually a two-step
process: First, data structures for neurons and synapses are instantiated
on the machine to define the network, including its connectivity;
afterwards, the network state is calculated over time starting from
some initial conditions according to the neuron and synapse dynamics.
This approach assumes that both neurons and their connectivity are
fixed once the network is constructed (typically using population-level
connection rules~\cite{Senk2022}) and at most the strength of connections
may be subject to change during the simulation~\cite{Morrison08_459}.
However, biological evidence suggests that the connectivity itself
is continuously changing, not only during development but also in
the adult brain~\cite{Fauth2016}. These changes are associated with
learning, adaptation, memory, recovery from lesions, drug addiction,
and occur following enriched experience, after sensory deprivation
or upon stimulation. The generation of new neurons in adult brains
(neurogenesis) is less ubiquitous, but has also been observed experimentally
-- in particular in the rodent hippocampal dentate gyrus~\cite{Altman1965,Kempermann2018}.
Besides the biological evidence, it has been found that algorithms
abstracted from structural plasticity can benefit machine-learning
(ML) models. Pruning~\cite{gale_state_2019} reduces the parameter
count of large models while preserving accuracy and, inspired by neurogenesis,
constructive neural networks~\cite{Frean1990,Kwok1997} are `grown'
as part of the training process. . The type of structural plasticity
addressed in this work does not add or remove neurons (as in most
brain regions beyond development) and concentrates on dynamic formation
and elimination of synapses.

To shed light on candidate structural plasticity mechanisms and their
function for learning and memory, computational models have been developed
to study, for example, whether changes are induced by activity, homeostasis,
stochasticity, or other factors. Simulator developers have implemented
functionality for modifying network connectivity during runtime to
simulate these models, e.g.~\cite{DiazPier2016,George2017,Bogdan2018,Liu2018,Rinke2018,Yan2019,Billaudelle2021}.
However, structural plasticity remains underexplored because simulating
networks with it is computationally expensive: First, structural changes
in biology are slow (hours, days) and therefore need long simulation
durations. Second, there are complex inter-dependencies with other
biological mechanisms such as synaptic plasticity and some processes
may need global instead of only local information, hence limiting
implementations. Third, continuously applying changes to the connectivity
clashes with the traditional approach of keeping the network structure
untouched in computer memory during state propagation which requires
to rethink how data structures are handled. Different simulation platforms
including neuromorphic systems therefore face different difficulties.

Structural plasticity algorithms have previously largely been implemented
in CPU-based SNN simulators~\cite[e.g.,][]{DiazPier2016,Vitay2015}
and neuromorphic systems such as SpiNNaker~\cite{Furber2014}, where
the main computational elements are scalar CPU cores~\cite[e.g.,][]{Bogdan2018,Liu2018}
and thus the standard routines used for modifying connectivity can
be employed. \textcite{Billaudelle2021} presented a structural plasticity
implementation for the BrainScaleS--2~ \cite{pehle_brainscales-2_2022}
Plasticity Processing Unit~(PPU) -- a Single Instruction Multiple
Thread~(SIMD) processor that is tightly integrated with the BrainScaleS--2
neuromorphic system. Here, we propose what we believe is the first
structural plasticity framework for implementation on general purpose
parallel hardware such as GPUs. The framework aims to combine flexibility
(i.e., the possibility to freely explore different structural plasticity
mechanisms) with efficiency (i.e., fast simulations). We identify
the following constraints:
\begin{enumerate}
\item Adding and removing connections cannot require the connectivity data
structures to be re-allocated.
\item The amount of data that needs to be moved around when adding or removing
a connection must be minimal.
\item The connectivity must still be indexed and read efficiently (i.e.,
via a coalesced memory read on a GPU).
\end{enumerate}
The GeNN simulator~\cite{Yavuz2016,Knight2021} was originally developed
for Computational Neuroscience research but its longstanding focus
on flexibility and its efficient implementation of sparse data structures
on GPU accelerators has enabled it to become a powerful tool for many
areas of SNN research. We have previously added support for various
machine-learning-specific functionalities including parallel batch
simulation of models -- allowing multiple copies of the model to
be run simultaneously in order to maximize GPU occupancy -- and user-defined
`custom update' operations~\cite{Nowotny2022,Knight2022}. Here,
we present a new extension to GeNN which allows users to easily define
hybrid CPU/GPU structural plasticity algorithms using the same code-generation
based approach already used to define neuron models and learning rules
as well as initialization algorithms.

Using this framework we demonstrate two applications. Firstly, building
on our previous work~\cite{Knight2022,Knight2023} using the e-prop
learning rule~\cite{Bellec2020} to efficiently train recurrent SNN
classifiers, we combine the GeNN implementation of e-prop with the
DEEP R structural plasticity rule~\cite{Bellec2018} -- implemented
in our new framework -- to train very sparse networks on the N-MNIST~\cite{orchard_converting_2015}
and DVS gesture classification tasks~\cite{Amir2017} with no loss
of accuracy while using $90\times$ fewer parameters than comparable
recurrent dense SNN models. The time overhead of applying structural
plasticity is less than $\unit[1]{\%}$ and the sparse models used
are an order of magnitude faster to train than their dense counterparts.
Secondly, we implement a model of topographic map formation~\cite{Bamford2010}
in GeNN, motivated by a recent example on the SpiNNaker platform~\cite{Bogdan2018}.
The spatially embedded network employs an activity-independent formation
rule: new connections are drawn according to a Gaussian dependence
on the lateral distance between neurons. However, synapse elimination
is activity-dependent: Spike-Timing-Dependent Plasticity~(STDP)~\cite{Morrison08_459}
varies synaptic strengths and weak synapses are more likely to be
removed. The interplay of different plasticity mechanisms in the model
results in a refinement of receptive fields and an embedding of learnt
input structures in the network connectivity. Even though the large
part of the simulation time is here spent on connectivity updates,
the simulation is still faster than previous implementations \cite{Bogdan2018}
and enables network upscaling experiments to test the scaling performance.

\section{Methods}

\label{sec:methods}

\subsection{GeNN custom connectivity updates\label{subsec:custom_conn_updates}}

\begin{figure}[H]
\begin{centering}
\includegraphics{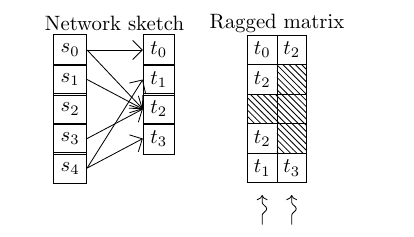}
\par\end{centering}
\caption{\textbf{Illustration of data structures used to implement sparse connectivity
in GeNN.} Five presynaptic neurons are connected to four postsynaptic
neurons using sparse connectivity with a maximum out-degree of two.
This is represented in memory as a \emph{ragged matrix}, where each
row n contains the indices of the target neurons of pre-synaptic neuron
n. Spikes are propagated through ragged matrix connectivity by parallelizing
across the columns as indicated by the snaking lines. \label{fig:ragged_matrix}}
\end{figure}

GeNN~\cite{Yavuz2016,Knight2021} is a library for generating CUDA~\cite{CUDA2020}
code for the simulation of SNN models. GeNN handles much of the complexity
of using CUDA and automatically performs device-specific optimizations
so as to maximize performance. GeNN consists of a main library --
implementing the API used to define models as well as the generic
parts of the code generator -- and an additional library for each
backend (currently there is a reference C++ backend for generating
CPU code and a CUDA backend, with an AMD HIP~\cite{advanced_micro_devices_amd_2021}
backend under development). PyGeNN~\cite{Knight2021} exposes these
libraries to Python using PyBind11~\cite{pybind11} and provides
a Pythonic and user-friendly interface to the low-level functionality
provided by GeNN.

One key advantage of GeNN over SNN simulation libraries such as SNNtorch~\cite{eshraghian2021training},
NORSE~\cite{norse2021} or JAXsnn~\cite{muller_jaxsnn_2024} --
which are built on standard Machine Learning toolkits like PyTorch
or JAX -- is that GeNN not only supports standard dense (i.e., all-to-all)
weight matrices stored as 2D arrays, but also sparse connectivity
stored in the `ragged matrix' data structure illustrated in \figref{ragged_matrix}.
In this data structure, each row contains the indices of a single
presynaptic neuron's postsynaptic targets and, to simplify address
calculation, rows are padded to the largest number of postsynaptic
targets present in the connectivity matrix. Additional 2D arrays with
the same shape are used to store synaptic variables such as weights
and delays. To efficiently parallelize the propagation of spikes through
this data structure, one CUDA thread is allocated to each column of
the matrix. Each thread then loops through incoming spikes, with each
`warp' of $32$ adjacent threads performing efficient coalesced reads
of synaptic weights and accumulating input to the postsynaptic neurons
using atomic operations (which on modern GPUs are handled efficiently
by the L2 cache). If connectivity updates can be parallelized across
presynaptic neurons, this format actually fits the constraints on
parallel connectivity updates we defined in the introduction rather
well. By adding additional padding to the end of each row, connections
can be added and removed without re-allocating the data structure.
Because the order of synapses in a row is unimportant, when adding
synapses they can simply be added to the end of the row and, when
removing them, the synapse at the end of the row can just be moved
to the location of the removed synapse and the out-degree reduced
by one.

Based on this foundation of efficient, parallel updating of connectivity
we reviewed the structural plasticity literature to determine the
types of functionality a general structural plasticity framework might
need to support.
\begin{description}
\item [{Access~to~synaptic~weights}] DEEP R~\cite{Bellec2018} needs
to detect changes to the sign of weights and the synaptic rewiring
rule developed by \textcite{Bamford2010} compares weights to a threshold.
\item [{Additional~pre~and~postsynaptic~variables}] DEEP R~\cite{Bellec2018}
needs to track the number of synapses which are removed from each
row of the synaptic matrix so they can be summed and the same number
of new synapses added.
\item [{Access~to~pre~and~postsynaptic~neuron~variables}] The rule
developed by \textcite{Butz2013} needs to access pre- and postsynaptic
calcium variables (essentially low-pass filtered spike trains).
\item [{Access~to~RNG}] DEEP R~\cite{Bellec2018} needs to be able to
sample from a random number generator to create new synapses and the
rule developed by \textcite{Bamford2010} uses probabilistic thresholds
to determine whether to actually eliminate or add candidate connections.
\end{description}
While, for convenience, GeNN provides some built-in models, one of
GeNN's most powerful features is that it enables users to easily define
their own models for neurons, synapses and custom updates (which are
typically used to implement optimizers and reductions for machine
learning models~\cite{Knight2023}) from within Python using strings
containing GeNNCode which is a subset of C99. Continuing this flexible
approach, we have added support for a new model primitive -- `custom
connectivity updates' -- to GeNN which provides a framework to build
structural plasticity rules. This framework is relatively complex
so, while in the remainder of this section we will try to give a flavor
of what is possible, we refer interested readers to the GeNN documentation
(\url{https://genn-team.github.io}) for more details. Rules can be
defined with the \texttt{pygenn.create\_custom\_connectivity\_update\_model}
helper function and can access neuron and synapse variables via `variable
references' (\texttt{pre\_var\_refs} and \texttt{post\_var\_refs};
and \texttt{var\_refs} respectively). Additionally, they provide one
GeNNCode string which is run on the GPU and parallelized across the
\emph{rows} of the synaptic matrix (\texttt{row\_update\_code}) and
another that is evaluated serially on the CPU (\texttt{host\_update\_code}).
Within all of these code strings, a suitable random number generator
is available via a family of \texttt{gennrand} functions. Within the
\texttt{row\_update\_code}, the index of the presynaptic neuron associated
with the row of connectivity can be accessed via the internal variable
\texttt{id\_pre} and synapses can be added using the \texttt{add\_synapse}
function which takes the index the postsynaptic target and the initial
values of any synaptic variables, in the order they are specified
in \texttt{var\_refs}. Variables which are associated with the connectivity
being updated but are not referenced in this way will be initialized
to zero for the new synapse. Using this functionality, we can define
a simple example which adds diagonal connections to the connectivity
matrix with a synaptic weight (referenced via \texttt{g}) of $1.0$:

\begin{minipage}{\linewidth}
\begin{lstlisting}[language=myPython]
add_diagonal_model = pygenn.create_custom_connectivity_update_model(
	"add_diagonal",             
	var_refs=[("g", "scalar")],
	row_update_code=                 
		"""
		add_synapse(id_pre, 1.0);
		""")
\end{lstlisting}
\end{minipage}

\noindent An instance of this custom connectivity update could then
be attached to an existing synapse group (\texttt{sg}) within a model
(\texttt{model}) and assigned to a \textit{`}custom update group\textit{'}
called ``update\_connectivity'':

\begin{minipage}{\linewidth}
\begin{lstlisting}[language=myPython]
cu = model.add_custom_connectivity_update(
    "add_diagonal", "update_connectivity", sg, add_diagonal_model,
    var_refs={"g": create_wu_var_ref(sg, "g")})
\end{lstlisting}
\end{minipage}

\noindent At runtime, the update can be triggered (along with any
other updates in the ``update\_connectivity'' group) by simply calling
\lstinline[language=myPython]{model.custom_update("update_connectivity")}.
If row updates need to remove or otherwise access existing synapses
they must iterate through them as the rows of the ragged matrix data
structure are unsorted. This is implemented by the \texttt{for\_each\_synapse}
statement, within the body of which, the index of current postsynaptic
targets can be accessed via the \texttt{id\_post} variable (as well
as the values of any variables or variable references associated with
individual synapses or postsynaptic neurons). Synapses can also be
removed within the body of the \texttt{for\_each\_synapse} statement
using the \texttt{remove\_synapse} function. Finally, custom connectivity
updates can define their own additional neuron and synapse variables
(\texttt{pre\_vars} and \texttt{post\_vars}; and \texttt{vars} respectively).
Using this additional functionality, we can define another example
which, on the CPU, first loops through each presynaptic neuron and
picks a random postsynaptic target to try and remove. Then, on the
GPU, we search each row of connectivity in parallel and, if the chosen
target is found, remove it:

\begin{minipage}{\linewidth}
\begin{lstlisting}[language=myPython]
remove_random_model = pygenn.create_custom_connectivity_update_model(
	"remove_random",
	pre_vars=[("postInd", "unsigned int")],
	row_update_code=
		"""
		for_each_synapse {
			if(id_post == postInd) {
				remove_synapse();
				break;
			}
		}
		""",
	host_update_code=
		"""
		for(unsigned int i = 0; i < num_pre; i++) {
			postInd[i] = genn_rand() % num_post;
		}                 
		pushpostIndToDevice();
		""")
\end{lstlisting}
\end{minipage}

\noindent In order to facilitate postsynaptic spike-triggered updates
to synapses, such as those involved in Spike Timing Dependent Plasticity~(STDP),
GeNN builds an additional transpose matrix data structure (see \cite{Knight2018}
for a complete description). This data structure is normally built
once at initialization time but, if custom connectivity updates are
applied to connections with these data structures, they get rebuilt
after each custom connectivity update.

GeNN provides timing functionality -- implemented on the GPU using
CUDA events and on the CPU using \texttt{std::chrono::high\_resolution\_clock}
-- to accurately measure the time spent in each simulation kernel
and we have extended this same system to enable the timing of the
CPU and GPU components of custom connectivity updates as well as of
the time spent `remapping' the transpose data structure.

\subsubsection{mlGeNN\label{subsec:mlGeNN}}

While we have previously implemented complex machine learning workflows
with networks of this sort in PyGeNN~\cite{Knight2022,Nowotny2022},
the syntax is designed for computational neuroscience and it can become
tedious and difficult to maintain. Therefore, we have built mlGeNN
on top of GeNN to provide a Keras-like abstraction~\cite{Turner2022,Knight2023}
with support for both converting pre-trained ANNs to SNNs and for
directly training SNNs using e-prop and Eventprop~\cite{wunderlich_event-based_2021}.
Here we have further extended mlGeNN to support DEEP R which can be
invoked by creating a sparsely-connected network and `compiling' it
using an\texttt{ EPropCompiler} instantiated as follows:

\begin{minipage}{\linewidth}
\begin{lstlisting}[language=myPython]
compiler = EPropCompiler(
    example_timesteps=num_timesteps, losses="sparse_categorical_crossentropy",
    optimiser="adam", batch_size=512,
	deep_r_conns=[in_hid, hid_hid], deep_r_l1_strength=0.005)
\end{lstlisting}
\end{minipage}

\subsubsection{DEEP R\label{subsec:Deep-Rewiring}}

\begin{figure}[H]
\begin{centering}
\includegraphics{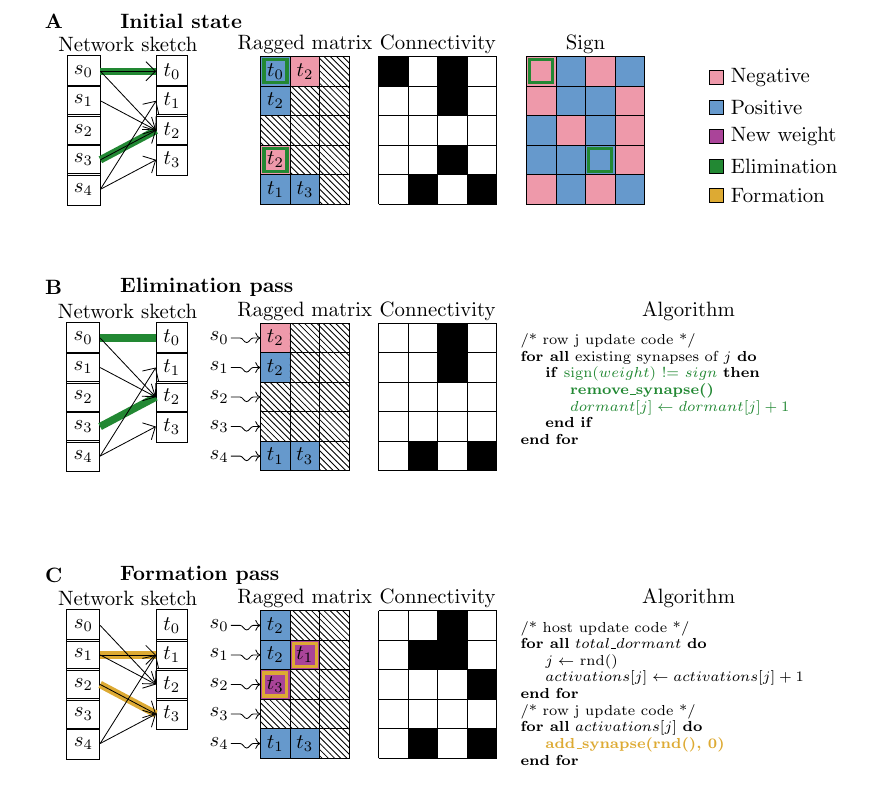}
\par\end{centering}
\caption{\textbf{Illustration of data structures used for the DEEP R implementation.}
Five presynaptic neurons are connected to four postsynaptic neurons
using sparse connectivity with a maximum out-degree of three. Each
panel shows a network sketch, the ragged matrix and two bitfields
-- one mirroring the connectivity to provide a faster lookup and
the other holding the fixed sign of each potential connection. \textbf{(A)}~Initial
state. Weights whose sign no longer matches the sign bit are highlighted
in green. \textbf{(B)}~Elimination pass. Connections highlighted
in (A) are removed, the number of connections in each row made `dormant'
are counted and the bitfield is updated. \textbf{(C)}~Formation pass.
Yellow highlighting indicates new the randomly added connections.\label{fig:deep_r}}
\end{figure}

The rewiring algorithm DEEP R~\cite{Bellec2018} is designed for
training efficient sparse neural networks by replacing weak connections
with randomly sampled new ones. DEEP R operates alongside a gradient-based
learning rule and marks synapses as `dormant' if weight updates would
have led to weights changing sign. Dormant connections are removed
and an equal number of randomly chosen synapses are added as replacements.
Unlike typical incremental pruning approaches, DEEP R maintains a
constant level of connection sparsity throughout training. However,
like many implementations of pruning in standard ML tools, the original
implementation represents the sparse connectivity as an additional
dense weight matrix containing the correct sign of each connection.
Here we implement a more efficient version using the structural connectivity
framework described in \subsecref{custom_conn_updates} which only
requires two additional bitfield data structures on top of the standard
GeNN sparse matrix format -- one to track the sign and one the presence
of each potential synaptic connection.

After the standard initialization of synaptic connectivity, a custom
connectivity update is used to initialize the two additional bitfields
as illustrated in \figref{deep_r}A. First, all connectivity bits
are cleared and all sign bits are randomized to fix the signs for
all \emph{potential} connections. Then, the update loops through all
the initialized synapses in each row, sets the corresponding connectivity
bit and ensures the corresponding sign bit matches the sign of the
initial weight of the \emph{actual} connection. After each training
batch, DEEP R is implemented by launching a sequence of custom updates.
\begin{description}
\item [{L1~regularisation}] is implemented by a custom update which adds
or subtracts (based on the sign bitfield) a constant from the gradients
accumulated by e-prop to push synaptic weights towards zero.
\item [{Elimination}] is implemented by a custom connectivity update which
loops through the synapses in each row and, if the sign of their weight
does not match the sign recorded in the bitfield, removes the synapses,
increments a per-row counter of `dormant' synapses and clears the
corresponding bit from the connectivity bitfield (\figref{deep_r}A--B).
\item [{Formation}] is implemented by a final custom connectivity update
which first uses the CPU to sum the row-wise dormant synapse count
and then re-distributes these uniformly as new `activations' between
the rows of synaptic connectivity. Then, on the GPU, these new synapses
are created for each activation, checking the connectivity bitfield
to ensure there are not duplicate synapses (\figref{deep_r}B--C).
\end{description}
In its original form~\cite{Bellec2018}, DEEP R also features an
additional noise term but, following \textcite{Bellec2020} -- who
first combined DEEP R with e-prop -- we do not include this in our
implementation.

\subsubsection{Synaptic rewiring for topographic maps\label{subsec:topomap_connectivity_update}}

\begin{figure}[H]
\begin{centering}
\includegraphics[width=150mm]{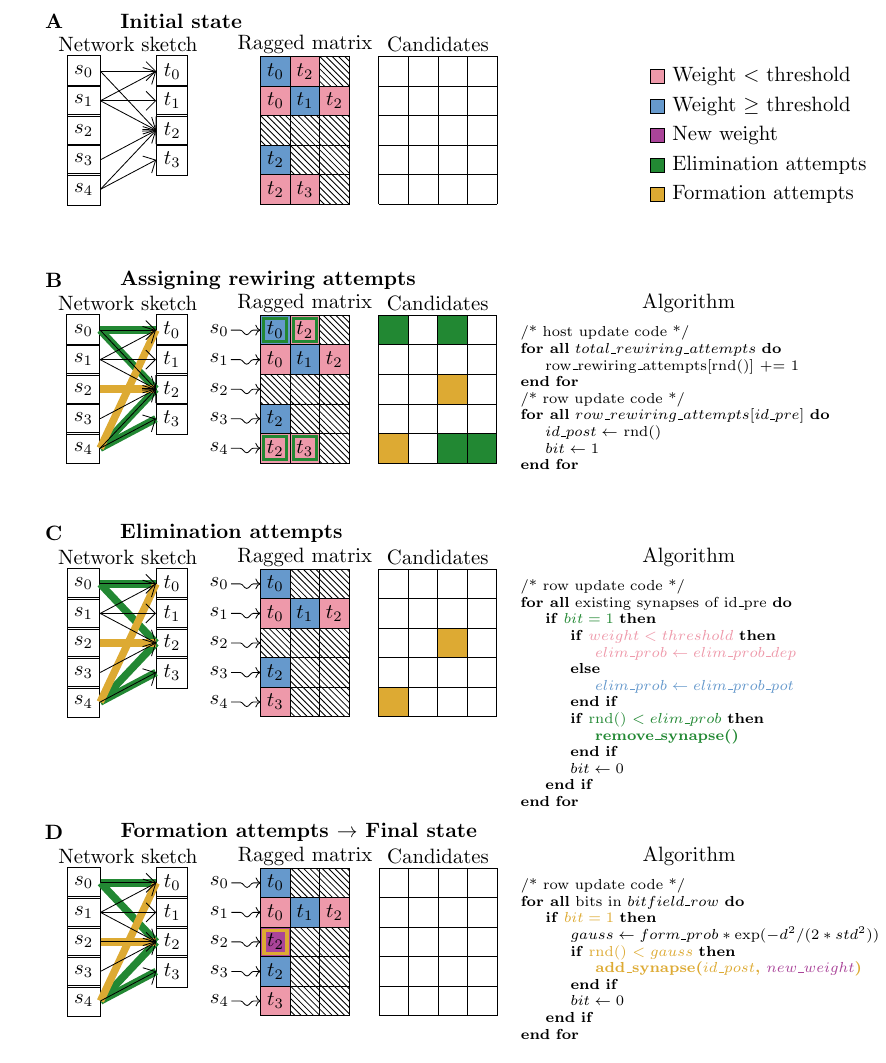}
\par\end{centering}
\caption{\textbf{Illustration of data structures used for one custom connectivity
update according to the synaptic rewiring model for topographic maps.}
Each panel shows a network sketch, the ragged matrix (number of presynaptic
neurons $\times$ rmaximum out-degree), the bitfield (number of presynaptic
neurons $\times$ number of postsynaptic neurons), and pseudocode.\textbf{
(A)}~Initial state. Five presynaptic neurons, sources $s_{j}$ with
$0\protect\leq j<5,$ are connected to four postsynaptic neurons,
targets $t_{i}$ with $0\protect\leq i<4$, using sparse connectivity
with a maximum out-degree of three. Weights below the threshold $g_{\theta}$
are indicated in light red, others in light blue. A synapse slot from
$s_{j}$ to $t_{i}$ is here referred to as $\left(s_{j},t_{i}\right)$.
The bitfield is cleared. \textbf{(B)}~Host assigns six rewiring attempts
randomly to three presynaptic neurons ($s_{0}$, $s_{2}$, $s_{4}$).
The presynaptic neurons then in parallel select their postsynaptic
targets randomly and set the corresponding bits as candidates for
rewiring attempts in the bitfield. Four existing synapses are selected
for elimination (green; $\left(s_{0},t_{0}\right)$, $\left(s_{0},t_{2}\right)$,
$\left(s_{4},t_{2}\right)$, $\left(s_{4},t_{3}\right)$) and two
empty synapse slots for formation (yellow; $\left(s_{2},t_{2}\right),\left(s_{4},t_{0}\right)$).
\textbf{(C)}~Elimination attempts. $\left(s_{0},t_{0}\right)$ has
a low elimination probability because of the strong weight (light
blue) and remains, the three others have a higher elimination probability
because of the lower weights (light red); $\left(s_{0},t_{2}\right)$
gets eliminated reducing the out-degree of $s_{0}$ in the ragged
matrix from two to one; $\left(s_{4},t_{2}\right)$ gets eliminated,
and the emerging gap in the row of $s_{4}$ is filled by inserting
the last element $t_{3}$; $\left(s_{4},t_{3}\right)$ remains. The
elimination bits get cleared in the bitfield. \textbf{(D)}~Formation
attempts. Assuming a spatial proximity of $s_{2}$ and $t_{2}$ and
therefore a high formation probability, a new synapse (purple) is
formed at $\left(s_{2},t_{2}\right)$, increasing the out-degree of
$s_{2}$ from zero to one, but not at $\left(s_{4},t_{0}\right)$
because $s_{4}$ and $t_{0}$ are farther apart. The formation bits
get cleared in the bitfield. \label{fig:topomap_datastructures}}
\end{figure}

We develop a GeNN-compatible variation of the synaptic rewiring model
proposed by \textcite{Bamford2010}, drawing inspiration from an implementation
on the SpiNNaker platform~\cite{Bogdan2018}. This stochastic model
combines a distance-dependent formation rule with a weight-dependent
removal rule. The simulation loop performs custom connectivity updates
at regular intervals according to a rewiring frequency. Each update
triggers a fixed number of rewiring attempts between neurons from
the presynaptic and postsynaptic population. While previous implementations~\cite{Bamford2010,Bogdan2018}
handle each rewiring attempt separately, our algorithm aims to increase
efficiency by parallelizing multiple attempts. \Figref{topomap_datastructures}
illustrates the hybrid CPU (host) and GPU (device) algorithm processing
these attempts and highlights the data structures. First, serial host
code (\texttt{host\_update\_code}) uniformly distributes the attempts
across all neurons in the presynaptic population. As we sample with
replacement, multiple attempts per presynaptic neuron are possible.
In contrast to earlier implementations~\cite{Bamford2010,Bogdan2018},
our version does not impose a fixed maximum in-degree. GeNN represents
connectivity in 2D-array structures; the rows correspond to presynaptic
neurons and can be operated on independently and hence in parallel.
Here, we introduce a bitfield as an auxiliary structure in a similar
format to mark synaptic slots between specific pre- and postsynaptic
neurons for wiring attempts. On the GPU (\texttt{row\_update\_code}),
each presynaptic neuron accesses a unique section of the bitfield
with as many bits as there are neurons in the postsynaptic population.
The bits therefore encode synapse slots for potential synapses between
the presynaptic neuron of the row considered and all postsynaptic
neurons; each bit refers to a postsynaptic neuron index. The bitfield
section is initialized to zero and then as many bits are set randomly
according to a uniform distribution without replacement as there are
wiring attempts assigned. Each synaptic slot is therefore considered
at most once, unlike the earlier versions~\cite{Bamford2010,Bogdan2018}
where each rewiring attempt is completely independent of the others.

When the presynaptic neuron has set its bitfield row, the elimination
rule is applied: A loop over all existing synapses of that presynaptic
neuron identifies those that are set in the bitfield section, removes
them dependent on an elimination probability compared to a random
number and clears the respective bit. Weak synapses are more likely
to be removed: the elimination probability is higher for depressed
synapses with weights below the threshold than for strong, potentiated
synapses. After all elimination attempts, the remaining set bits in
the bitfield row point to synapse slots that were initially empty.
Applying the formation rule, we then iterate over the whole bitfield
row: If a bit is still set, we attempt to establish a synapse according
to a Gaussian probability depending on the distance between the neurons
and also clear that bit. The shorter the lateral distance between
pre- and postsynaptic neuron, the more likely is a new connection.
Our interpretation of the rewiring model does not enable the formation
of multapses (i.e., multiple synapses per neuron pair~\cite{Senk2022}).
If a synapse exists, only the elimination rule is applied. To sum
up, each custom connectivity update initiates a fixed number of rewiring
attempts between pre- and postsynaptic neuron populations. The outcome
of each attempt is a newly formed connection, a removed connection,
or no change to the connectivity.

\subsection{Network models}

\subsubsection{Recurrent classifier\label{subsec:Recurrent-Classifier}}

Our previous work~\cite{Nowotny2022,Knight2022,Knight2023} showed
that recurrent SNN classifiers can be trained on datasets such as
the Spiking Heidelberg Digits~\cite{Cramer2020} and DVS gesture~\cite{Amir2017}.

Here we train SNN classifiers with a single recurrently connected
hidden layers of 512 neurons on the N-MNIST and DVS gesture datasets.
Following the approach outlined by \textcite{subramoney2023efficient},
we spatially downsample the DVS gesture examples from $128\times128$
to $32\times32$ resolution and temporally downsample so that there
is only one spike in each $\unit[1]{ms}$ simulation timestep. When
training on DVS gesture, we use adaptive LIF~(ALIF) neurons, the
membrane voltage ($v$), adaptation variable ($a$) and spiking output
($z$) of which can be described by:

\begin{align}
v_{j}^{t+1} & =\alpha\left(v_{j}^{t}-z_{j}^{t}v_{\textrm{thr}}\right)+\sum_{i\neq j}W_{ji}^{\textrm{rec}}z_{i}^{t}+\sum_{i}W_{ji}^{\textrm{in}}x_{i}^{t}\\
a_{j}^{t+1} & =\rho a_{j}^{t}+z_{j}^{t}\\
z_{j}^{t} & =H(v_{j}^{t}-(v_{\textrm{thr}}+\beta a_{j}^{t})),
\end{align}
where $\alpha=e^{-\frac{\textrm{dt}}{\tau_{\textrm{mem}}}}$represents
the decay of the membrane with time constant $\tau_{\textrm{mem}}=\unit[20]{ms}$,
$W^{\textrm{rec}}$ are the recurrent weights, $W^{\textrm{in}}$
are the weights connecting the input to the recurrent layer, $x^{t}$
are input spikes from the dataset, $v_{\textrm{thr}}=0.6$ is the
baseline spiking threshold, $\rho=e^{-\frac{\textrm{dt}}{\tau_{\textrm{adapt}}}}$represents
the adaptation variable decay with time constant $\tau_{\textrm{adapt}}=\unit[2000]{ms}$
and $\beta=0.0174$ is the adaptation strength. When neurons spike,
$v_{\textrm{thr}}$ is subtracted from their membrane voltage $v$.
In our classifiers, $W^{\textrm{rec}}$ and $W^{\textrm{in}}$ are
sparse and initialised with a fixed probability of connection between
each pair of pre and postsynaptic neurons e.g. we use $\unit[5]{\%}$
connectivity to indicate that there is a $\unit[5]{\%}$ chance of
a pair of pre and postsynaptic neurons being connected. When training
on the simpler N-MNIST task, we use standard LIF neurons, the dynamics
of which can be obtained by setting $\beta=0$ and $\rho=0$.

The network readout ($y$) is implemented using simple leaky integrators:

\begin{align}
y_{k}^{t+1} & =\alpha y_{k}^{t}+\sum_{i}W_{ki}^{\textrm{out}}z_{j}^{t}+b_{k}^{\textrm{out}},
\end{align}
where $W^{\textrm{out}}$ represents the weights connecting the hidden
layer to the output and $b^{\textrm{out}}$ are learned biases.

We calculate weight gradients $\Delta W^{\textrm{rec}}$ and $\Delta W^{\textrm{in}}$
using eligibility traces $\epsilon$ and $e$ calculated using e-prop~\cite{Bellec2020}
(shown here for $\Delta W^{\textrm{rec}}$) :

\begin{align}
\epsilon_{ji}^{t+1} & =\psi_{j}^{t}\bar{z}_{i}^{t}+(\rho-\psi_{j}^{t}\beta)\epsilon_{ji}^{t}\label{eq:eprop_elig_1}\\
e_{ji}^{t} & =\psi_{j}^{t}(\bar{z}_{i}^{t}-\beta)\epsilon_{ji}^{t}\label{eq:eprop_elig_2}\\
\Delta W_{ji}^{\textrm{rec}} & =\sum_{t}\left(\sum_{k}B_{jk}(\pi_{k}^{t}-\pi_{k}^{*,t})\right)\bar{e}_{ji}^{t},
\end{align}
where $\psi_{j}^{t}=\frac{0.3}{v_{\textrm{thr}}}\text{max}\left(0,1-\left|\frac{v_{j}^{t}-(v_{\textrm{thr}}+\beta a_{j}^{t})}{v_{\textrm{thr}}}\right|\right)$
is a surrogate gradient function, $\bar{z}^{t}$ is a version of $z^{t}$
low-pass filtered with a time constant of $\tau_{\textrm{mem}}$,
$\eta$ is a learning rate, $B=\left(W^{\textrm{out}}\right)^{T}$,
$\pi_{k}^{t}=\textrm{softmax}(y_{1}^{t},\ldots,y_{K}^{t})$, $\pi^{*,t}$
are one-hot encoded labels and $\bar{e}^{t}$ is a version of $e^{t}$
low-pass filtered with a time constant of $\tau_{\textrm{mem}}$.
When using LIF neurons, only a single simpler eligbility trace is
required:

\begin{align}
e_{ji}^{t+1} & =\psi_{j}^{t+1}\bar{z}_{i}^{t}\label{eq:eprop_elig_lif}
\end{align}

\noindent Finally, output weight gradients ($\Delta W^{\textrm{out}}$)
and biases ($\Delta b^{\textrm{out}}$) are calculated using a simple
delta rule:

\begin{align}
\Delta W_{kj}^{\textrm{out}} & =\sum_{t}(\pi_{k}^{t}-\pi_{k}^{*,t})\bar{z}_{j}^{t}\\
\Delta b_{k}^{\textrm{out}} & =\sum_{t}(\pi_{k}^{t}-\pi_{k}^{*,t}).
\end{align}

\noindent All gradients are applied to the underlying variables using
the Adam optimizer~\cite{Kingma2015} and $\Delta W^{\textrm{rec}}$
and $\Delta W^{\textrm{in}}$provide the inputs to the DEEP R algorithm
described in \subsecref{Deep-Rewiring}.

\subsubsection{Topographic map}

\begin{figure}[H]
\centering{}\includegraphics[width=15cm]{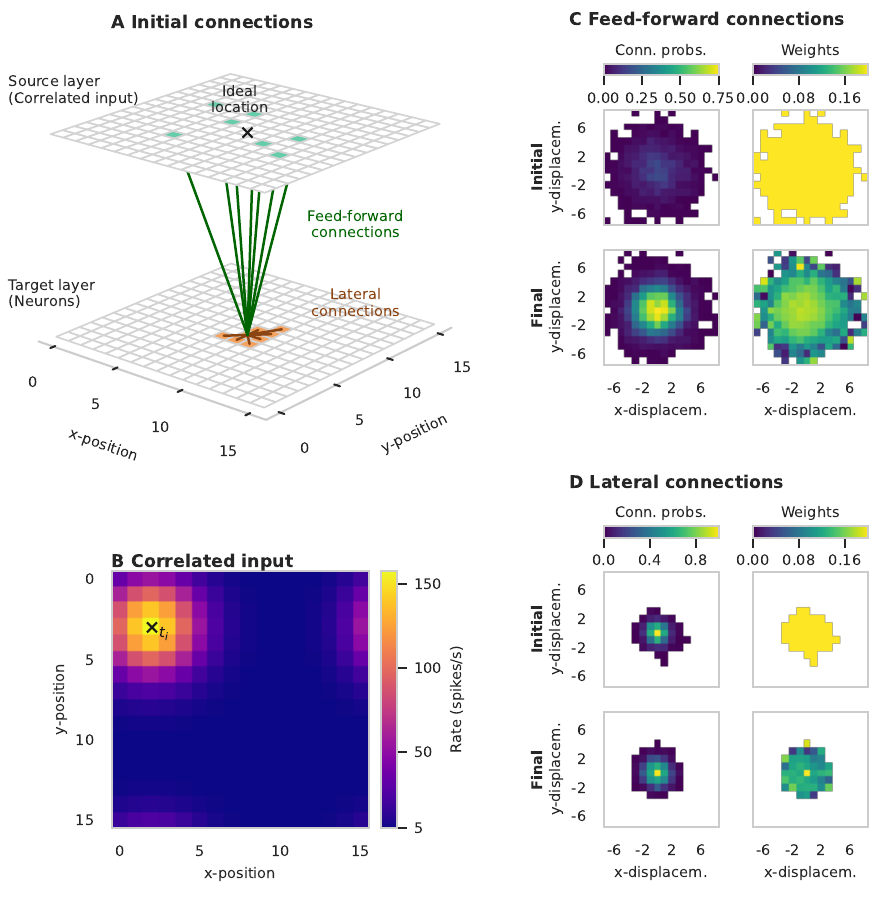}\caption{\textbf{Topographic map model. (A)}~Network sketch indicating initial
connections and ideal location of the preferred location for one neuron.\textbf{
(B)}~Spatially correlated input firing rates; the center (x-marker)
changes in regular time intervals. Connection probabilities and weights
for \textbf{(C)}~feed-forward and\textbf{ (D)}~lateral connections,
averaged over displacements for each grid position. \label{fig:topomap_methods}}
\end{figure}

\begin{table}[H]
\begin{tabular}{|>{\raggedright}p{3cm}|>{\raggedright}p{11.5cm}|}
\hline 
\multicolumn{2}{|l|}{\cellcolor{black}\color{white}{\textbf{A~Model summary}}}\tabularnewline
\hline 
\textbf{Populations} & Two: source layer (Poisson inputs), target layer (excitatory neurons)\tabularnewline
\hline 
\textbf{Topology} & Nodes positioned equally spaced on square grid domain, periodic boundary
conditions\tabularnewline
\hline 
\textbf{Connectivity} & Feed-forward and lateral connections, initialization with spatial
profile\tabularnewline
\hline 
\textbf{Neuron} & Conductance-based leaky integrate-and-fire (LIF)\tabularnewline
\hline 
\textbf{Input} & Poisson spike trains with spatially correlated rate\tabularnewline
\hline 
\textbf{Plasticity} & Synaptic plasticity (Spike-Timing-Dependent Plasticity (STDP)), structral
plasticity\tabularnewline
\hline 
\end{tabular}

\begin{tabular}{|>{\raggedright}p{3cm}|>{\raggedright}p{11.5cm}|}
\hline 
\multicolumn{2}{|l|}{\cellcolor{black}\color{white}{\textbf{B~Neuron, input and plasticity
models}}}\tabularnewline
\hline 
\textbf{Neuron\newline (target layer)} & Conductance-based leaky integrate-and-fire (LIF)

\begin{itemize} \itemsep0em 

\item  Membrane potential $V_{i}\left(t\right)$ of neuron $i$:

\item[] Spike emission at times $t_{s}^{i}\in\mathcal{S}_{i}=\left\{ t_{0}^{i},t_{1}^{i},t_{2}^{i},...\right\} $
with $V_{i}\left(t_{s}^{i}\geq V_{\theta}\right)$ \item[]  Subthreshold
dynamics: $\tau_{\text{m}}\frac{\text{d}V_{i}}{\text{d}t}=V_{\text{rest}}-V_{i}+\frac{g_{i}}{g_{\text{leak}}}\left(E_{\text{exc}}-V_{i}\right)\qquad\text{with}\qquad g_{\text{leak}}=\frac{C_{\text{m}}}{\tau_{\text{m}}}$\newline 
if $\forall s:t\notin\left(t_{s}^{i},t_{s}^{i}+\tau_{\text{ref}}\right]$

\item[]  Reset and refractoriness: $V_{i}\left(t\right)=V_{\text{reset}}$
if $t\in\left(t_{s}^{i},t_{s}^{i}+\tau_{\text{ref}}\right],t_{s}^{i}\in\mathcal{\mathcal{S}}_{i}$
\item Conductance of neuron $i$ with presynaptic neurons $j$:

$\tau_{\mathrm{s}}\frac{\text{d}g_{i}}{\text{d}t}=-g_{i}+\sum_{j}g_{\text{syn},ij}\sum_{s}\delta\left(t-t_{s}^{j}-t_{d}\right)$\end{itemize}\tabularnewline
\hline 
\textbf{Input\newline (source layer)} & Poisson spike trains

\begin{itemize} \itemsep0em \item Spatially correlated firing rate
$f\left(i\right)=f_{\text{base}}+f_{\text{peak}}e^{-\frac{d_{l,i}^{2}}{2\sigma_{\text{stim}}^{2}}}$
\newline with Euclidean distance $d_{il}$ between stimulation center
$l$ and grid location $i$\item  Stimulation center changes randomly
after each interval $t_{\text{stim}}$\end{itemize}\tabularnewline
\hline 
\textbf{Synaptic plasticity} & Spike-Timing-Dependent Plasticity (STDP) \begin{itemize}  \item 
All-to-all spike pairing for presynaptic neuron $j$ and postsynaptic
neuron $i$

$\frac{\text{d}g_{\text{syn},ij}}{\text{d}t}=A_{+}x_{j}\delta\left(t-t_{s}^{i}\right)-A_{-}y_{i}\delta\left(t-t_{s}^{j}\right)$
with traces\newline $-\frac{x_{j}}{\tau_{+}}+\sum_{s}\delta\left(t-t_{s}^{j}\right)$
and $\frac{\text{d}y_{i}}{\text{d}t}=-\frac{y_{i}}{\tau_{-}}+\sum_{s}\delta\left(t-t_{s}^{i}\right)$

\item Conductances restricted to $0\leq g_{\text{syn}}\leq g_{\text{max}}$

\end{itemize}\tabularnewline
\hline 
\textbf{Structural plasticity} & Probabilistic connection changes with random number $R$ sampled from
uniform distribution $[0,1)$ \begin{itemize} \itemsep0em \item 
Activity-dependent elimination:

\[
R<p_{\text{elim}}\,\text{where}\,p_{\text{elim}}=\begin{cases}
p_{\text{elim\_dep}} & \text{for}\,g_{\text{syn}}<g_{\theta}\\
p_{\text{elim\_pot}} & \text{otherwise}
\end{cases}
\]
\item  Distance-dependent formation: 
\[
R<p_{\text{form}}\text{e}^{-\frac{d_{ij}^{2}}{2\sigma_{\text{form}}^{2}}}
\]
with Euclidean distance $d_{ij}$ between grid locations $i$ and
$j$, new synapse initialized with $g_{\text{max}}$ \item  $N_{\text{attempts}}$
rewiring attempts after each interval $t_{\text{rewiring}}$ \end{itemize}\tabularnewline
\hline 
\end{tabular}

\caption{Topographic map model description and parameters following the guidelines
of \textcite{Nordlie2009}. \label{tab:topomap_model_description}}
\end{table}

\begin{table}[H]
\begin{tabular}{|>{\raggedright}p{2cm}|>{\raggedright}p{2cm}|>{\raggedright}p{10cm}|}
\hline 
\multicolumn{3}{|l|}{\cellcolor{black}\color{white}{\textbf{C~Topology}}}\tabularnewline
\hline 
\textbf{Symbol} & \textbf{Value} & \textbf{Description}\tabularnewline
\hline 
$L_{0}$ & $16$ & Unitless domain length at scale $s=0$\tabularnewline
\hline 
$N_{0}$ & $L_{0}\times L_{0}=256$ & Network size (number of nodes per layer) at scale $s=0$\tabularnewline
\hline 
$s$ & $\left\{ 0,...,7\right\} $ & Scaling factor for domain length\tabularnewline
\hline 
\end{tabular}
\begin{raggedright}
\begin{tabular}{|>{\raggedright}p{2cm}|>{\raggedright}p{2cm}|>{\raggedright}p{10cm}|}
\hline 
\multicolumn{3}{|l|}{\cellcolor{black}\color{white}{\textbf{D~Neuron}}}\tabularnewline
\hline 
\textbf{Symbol} & \textbf{Value} & \textbf{Description}\tabularnewline
\hline 
$C_{\mathrm{m}}$ & $\unit[20]{nF}$ & Membrane capacitance\tabularnewline
\hline 
$\tau_{\mathrm{m}}$ & $\unit[20]{ms}$ & Membrane time constant\tabularnewline
\hline 
$V_{\text{rest}}$ & $\unit[-70]{mV}$ & Resting potential\tabularnewline
\hline 
$E_{\text{ext}}$ & $\unit[0]{mV}$ & Reversal potential\tabularnewline
\hline 
$V_{\theta}$ & $\unit[-54]{mV}$ & Firing threshold\tabularnewline
\hline 
$V_{\mathrm{reset}}$ & $\unit[-70]{mV}$ & Reset potential\tabularnewline
\hline 
$\tau_{\mathrm{ref}}$ & $\unit[5]{ms}$ & Absolute refractory period\tabularnewline
\hline 
$\tau_{s}$ & $\unit[5]{ms}$ & Synaptic time constant\tabularnewline
\hline 
$t_{d}$ & $\unit[0.1]{ms}$ & Spike transmission delay\tabularnewline
\hline 
\end{tabular}
\par\end{raggedright}
\begin{raggedright}
\begin{tabular}{|>{\raggedright}p{2cm}|>{\raggedright}p{2cm}|>{\raggedright}p{10cm}|}
\hline 
\multicolumn{3}{|l|}{\cellcolor{black}\color{white}{\textbf{E~Input}}}\tabularnewline
\hline 
\textbf{Symbol} & \textbf{Value} & \textbf{Description}\tabularnewline
\hline 
$f_{\text{base}}$ & $\unit[5]{Hz}$ & Base firing rate\tabularnewline
\hline 
$f_{\text{peak}}$ & $\unit[152.8]{Hz}$ & Peak firing rate\tabularnewline
\hline 
$\sigma_{\text{stim}}$ & $2$ & Standard deviation of stimulus spread\tabularnewline
\hline 
$t_{\text{stim}}$ & $\unit[20]{ms}$ & Interval for changing stimulation center\tabularnewline
\hline 
\end{tabular}
\par\end{raggedright}
\begin{raggedright}
\begin{tabular}{|>{\raggedright}p{2cm}|>{\raggedright}p{2cm}|>{\raggedright}p{10cm}|}
\hline 
\multicolumn{3}{|l|}{\cellcolor{black}\color{white}{\textbf{F~Synaptic plasticity}}}\tabularnewline
\hline 
\textbf{Symbol} & \textbf{Value} & \textbf{Description}\tabularnewline
\hline 
$g_{\text{max}}$ & $\unit[0.2]{mS}$ & Maximum conductance\tabularnewline
\hline 
$w_{\text{min}}$ & $\unit[0]{mS}$ & Minimum allowed conductance\tabularnewline
\hline 
$w_{\text{max}}$ & $g_{\text{max}}$ & Maximum allowed conductance\tabularnewline
\hline 
$B$ & $1.2$ & Ratio of potentiation to depression es defined in \cite{Song2001}\tabularnewline
\hline 
$A_{+}$ & $0.1\times g_{\text{max}}$ & Learning rate for potentiation\tabularnewline
\hline 
$\tau_{+}$ & $\unit[20]{ms}$ & Time constant for potentiation\tabularnewline
\hline 
$\tau_{-}$ & $\unit[64]{ms}$ & Time constant for depression\tabularnewline
\hline 
$A_{-}$ & $BA_{+}\frac{\tau_{+}}{\tau_{-}}$ & Learning rate for depression\tabularnewline
\hline 
\end{tabular}
\par\end{raggedright}
\begin{raggedright}
\begin{tabular}{|>{\raggedright}p{2cm}|>{\raggedright}p{2cm}|>{\raggedright}p{10cm}|}
\hline 
\multicolumn{3}{|l|}{\cellcolor{black}\color{white}{\textbf{G~Structural plasticity}}}\tabularnewline
\hline 
\textbf{Symbol} & \textbf{Value} & \textbf{Description}\tabularnewline
\hline 
$p_{\text{form-ff}}$ & $0.16$ & Peak formation probability for feed-forward connections\tabularnewline
\hline 
$\sigma_{\text{form-ff}}$ & $2.5$ & Standard deviation of receptive field for feed-forward connections\tabularnewline
\hline 
$p_{\text{form-lat}}$ & $1$ & Peak formation probability for lateral connections\tabularnewline
\hline 
$\sigma_{\text{form-lat}}$ & $1$ & Standard deviation of receptive field for lateral connections\tabularnewline
\hline 
$g_{\theta}$ & $\frac{1}{2}g_{\text{max}}$ & Conductance threshold\tabularnewline
\hline 
$p_{\text{elim-dep}}$ & $2.45\times10^{-2}$\newline $\times50$ & Elimination probability for depressed synapses ($g_{\text{syn}}<g_{\theta}$);
value from \cite{Bamford2010} increased by factor\tabularnewline
\hline 
$p_{\text{elim-pot}}$ & $1.36\times10^{-4}$\newline $\times50$ & Elimination probability for potentiated synapses ($g_{\text{syn}}\geq g_{\theta}$);
value from \cite{Bamford2010} increased by factor\tabularnewline
\hline 
$t_{\text{rewiring}}$ & $\unit[1]{ms}$ & Rewiring interval\tabularnewline
\hline 
$N_{\text{attempts}}$ & $10$ & Number of rewiring attempts after each rewiring interval\tabularnewline
\hline 
\end{tabular}
\par\end{raggedright}
\begin{raggedright}
\begin{tabular}{|>{\raggedright}p{2cm}|>{\raggedright}p{2cm}|>{\raggedright}p{10cm}|}
\hline 
\multicolumn{3}{|l|}{\cellcolor{black}\color{white}{\textbf{H~Simulation}}}\tabularnewline
\hline 
\textbf{Symbol} & \textbf{Value} & \textbf{Description}\tabularnewline
\hline 
$T_{\text{model}}$ & $\unit[60]{s}$ & Biological model time\tabularnewline
\hline 
$h$ & $\unit[0.1]{ms}$ & Simulation time step\tabularnewline
\hline 
\end{tabular}
\par\end{raggedright}
\caption{Continuation of \tabref{topomap_model_description}. \label{tab:topomap_parameters}}
\end{table}

The network model illustrated in \figref{topomap_methods}A and its
parameters closely match the earlier accounts on topographic map formation
of \textcite{Bamford2010,Bogdan2018}. In this section, we provide
a summary and emphasize the differences, while tables \ref{tab:topomap_model_description}
and \ref{tab:topomap_parameters} give a complete overview of the
model description and parameters used in our work in the format frequently
used in computational neuroscience \cite{Nordlie2009}. The model
consists of two square layers with nodes located at grid positions:
model neurons in the target layer and spike-generating devices in
the source layer. Feed-forward connections are established from the
source to the target layer and lateral connections within the target
layer. Our work is inspired by ``Case 1: STDP and rewiring operating
simultaneously in the presence of correlated input'' of~\cite{Bogdan2018}.
The process of creating a topographic mapping is driven by correlated
input from the devices in the source layer which generate spikes with
Poisson-distributed inter-spike intervals according to spatially modulated
firing rates following a Gaussian decay from a stimulus center that
changes location randomly in regular intervals as illustrated in \figref{topomap_methods}B.
After network initialization, both feed-forward and lateral connections
are continuously updated with synaptic plasticity (i.e., weight adjustments
according to an STDP rule) and structural plasticity (i.e., changes
to the connectivity itself as described in \subsecref{topomap_connectivity_update}
in line with the ``fast'' rewiring protocol of~\cite{Bogdan2018}).
Starting from a rough topographic map, simulating the model leads
to a refinement of receptive fields: each neuron's preferred location
approximates its ideal location. Note that the formation rule assumes
wider feed-forward than lateral connections as shown in \figref{topomap_methods}C
and D. We deviate from previous implementations \cite{Bamford2010,Bogdan2018}
in the following points:
\begin{itemize}
\item Since SpiNNaker internally scales with a maximum weight, we multiply
the STDP learning rate $A_{+}$ given in their table~1 with their
$g_{\text{max}}$ to get the same effect.
\item Instead of nearest-neighbor spike pairing, we use all-to-all pairing
of pre- and postsynaptic spikes to compute the weight updates of the
STDP model~\cite{Morrison08_459}.
\item We also use the formation rule to initialize the connectivity, but
instead of fixing the initial in-degree, we apply the ``pairwise Bernoulli''
rule~\cite{Senk2022} using probabilities computed according to the
formation probability given in \tabref{topomap_model_description}.
This approach leads to on average $K=2\pi p_{\text{form}}\sigma_{\text{form}}$
synapses per neuron instead of $32$ and excludes multapses~\cite{Senk2022}
from the beginning as these are not supported by the structural plasticity
rule.
\item To keep the formation and elimination in balance without fixing the
maximum in-degree, we increase both of the elimination probabilities
$p_{\text{elim\_dep}}$ and $p_{\text{elim\_pot}}$ by a factor of
$50$.
\item Our rewiring attempts are executed in parallel for each presynaptic
neuron instead of sequentially.
\item We use a simulation time step of $\unit[h=0.1]{ms}$ (same as \cite{Bamford2010})
in contrast to $\unit[1]{ms}$ \cite{Bogdan2018} and solve the neuron
dynamics with the exponential Euler method. Our delay also corresponds
to one time step.
\end{itemize}
Going beyond the network size used in the original model, we scale
the domain length $L_{0}$ with an integer scaling factor $s$. To
preserve the dynamics, we replicate the correlated input in each added
square such that there are $s^{2}$ stimulus locations selected simultaneously
in each stimulation interval. The number of rewiring attempts in
the full network is also scaled with $s^{2}$. For inspecting the
connectivity evolution, we read out the connectivity after each connectivity
update which slows down the simulation; for performance measurements
we do not do such readouts.

\subsection{Software and hardware summary}

Topographic map model simulations are performed on one GPU of an NVIDIA
DGX A100 system using GeNN 5.3.0 and Python 3.12.3. DEEP R training
was performed on the JUWELS Booster HPC system, using an environment
with Python 3.11.3, GeNN 5.2.0 and mlGeNN 2.4.0. All analysis and
plotting uses Python with numpy 1.26.4, and matplotlib 3.8.4. All
source codes to reproduce our results are publicly available on GitHub
(\url{https://github.com/jhnnsnk/genn_structural_plasticity}).

\section{Results}

\label{sec:results}

\subsection{Classification with DEEP R}

\begin{figure}[H]
\begin{centering}
\includegraphics{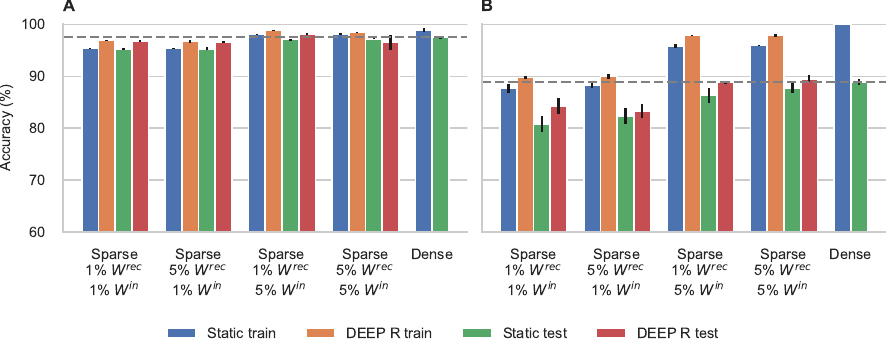}
\par\end{centering}
\caption{\textbf{Classification accuracy.} Bar heights show mean accuracy and
error bars standard deviations, all calculated over five repeats with
different seeds. The dashed line shows the mean test accuracy of the
baseline dense network.\textbf{ (A)}~N-MNIST. \textbf{(B)~}DVS-gesture.\label{fig:deep_r_accuracy}}
\end{figure}

\begin{figure}[H]
\begin{centering}
\includegraphics{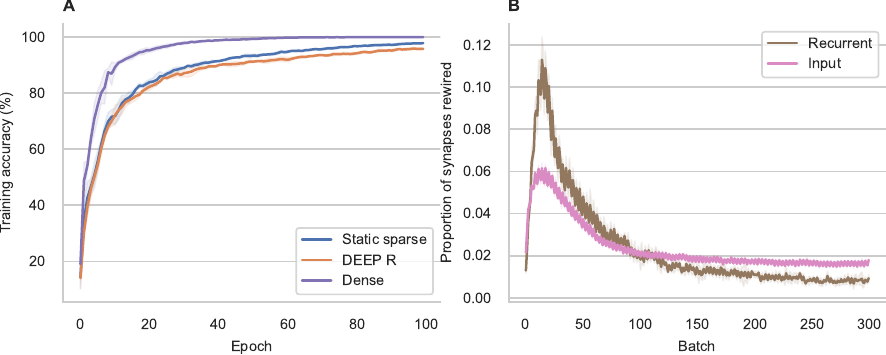}
\par\end{centering}
\caption{\textbf{DVS gesture training dynamics.} Lines show mean values and
shaded areas indicate one standard deviation, all calculated over
five repeats with different seeds. Sparse models have $\unit[5]{\%}$
$W^{\textrm{in}}$ connectivity and $\unit[1]{\%}$ $W^{\textrm{rec}}$
connectivity.\textbf{ (A)}~Training convergence. \textbf{(B)}~DEEP
R rewiring. Proportion of synapses rewired over the course of training.\label{fig:DVS-gesture-training}}
\end{figure}

\begin{figure}[H]
\begin{centering}
\includegraphics{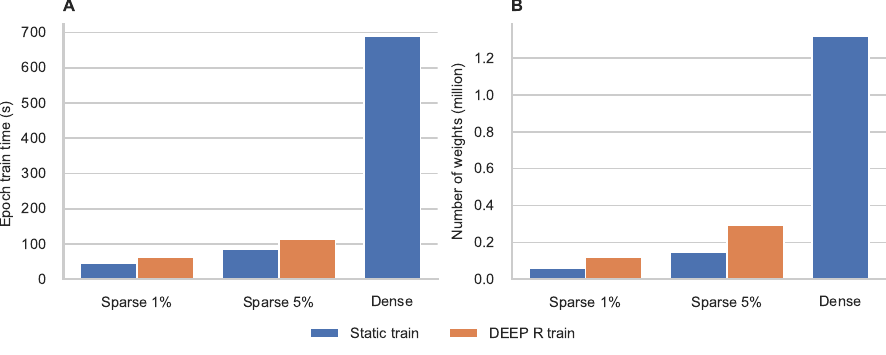}
\par\end{centering}
\caption{\textbf{Performance and weight savings of sparse DVS gesture training.}
Sparse models shown have the same sparsity in $W^{\textrm{in}}$and
$W^{\textrm{rec}}$.\textbf{ (A)~}Training time.\textbf{(B)~}The
number of weights in models.\label{fig:deep_r_perf_mem}}
\end{figure}

Figure~\ref{fig:deep_r_accuracy} shows the training and testing
accuracy of sparse and dense classifier networks (\subsecref{Recurrent-Classifier})
trained on the N-MNIST~\cite{orchard_converting_2015} and DVS gesture
datasest~\cite{Amir2017} using mlGeNN~(see \subsecref{mlGeNN})
with the GPU-parallelised DEEP R implementation described in \subsecref{Deep-Rewiring}.
We trained all models for $100$ epochs using a batch size of $512$
and the Adam optimizer~\cite{Kingma2015} with default parameters.
We also used the default e-prop parameters for all networks  only
optimizing the L1 regularization strength on a validation set. Neither
dataset contains a dedicated validation split so, as N-MNIST has a
large number of examples, we simply held out 10000 random example
whereas, for DVS gesture, we employed a cross-validation methodology
-- where we leave out two users in each fold (after \textcite{Nowotny2022}).

Except for the model trained on DVS gesture with $\unit[5]{\%}$ input
connectivity and $\unit[1]{\%}$ recurrent connectivity; and the model
trained on N-MNIST with $\unit[5]{\%}$ input connectivity and $\unit[5]{\%}$
recurrent connectivity, using DEEP R significantly improves the performance
of all sparse models compared to the versions trained with fixed connectivity
($p<0.05$, paired, one-sided Wilcoxon signed rank test~\cite{wilcoxon1945individual},
$n=5$). Furthermore, as long as there are sufficiently many input
connections ($\geq$$\unit[5]{\%}$ input connectivity), sparse models
trained with DEEP R matched the performance of the dense reference
networks on both datasets. Specifically, our models with $\unit[5]{\%}$
input connectivity and $\unit[1]{\%}$ recurrent connectivity trained
on DVS gesture obtains a test accuracy of $\unit[88.86\pm0.3]{\%}$
compared to $\unit[88.94\pm0.6]{\%}$ for the dense model. This is
higher than the $\unit[88.0]{\%}$ reported by \textcite{subramoney2023efficient}
for an event-based Gated Recurrent Unit~(GRU) model and is achieved
with around $90\times$ fewer parameters ($60\times10^{3}$ vs. $5.5\times10^{6}$).

\Figref{DVS-gesture-training}A shows the training dynamics of three
types of models on the DVS gesture dataset and demonstrates that the
training of all types is stable, although, as the literature predicts,
the sparser models train more slowly~\cite{frankle_lottery_2019}.
Figure~\ref{fig:DVS-gesture-training}B shows how the rate of DEEP
R rewiring changes over time with drastic changes in connectivity
happening during the first $100$ batches but connectivity remaining
more or less unchanged towards the end of training, matching the intuition
that DEEP R is sampling network connectivity and then settling into
a possible solution.

We then measured computational speed of DVS gesture training. As shown
in \figref{deep_r_perf_mem}A, training models with $\unit[5]{\%}$
connectivity using e-prop and GeNN is up to $8\times$ faster than
training a dense model and training models with $\unit[1]{\%}$ connectivity
is up to $15\times$ faster. While the time spent running the DEEP
R updates described in \subsecref{Deep-Rewiring} takes a maximum
of $\unit[0.5]{\%}$ of the total training time, training the models
with DEEP R is slightly slower because we (conservatively) double
the maximum out-degree in the ragged sparse matrix data structure
(see \subsecref{custom_conn_updates}) to allow space for rewiring.
\Figref{deep_r_perf_mem}B shows the number of weights in each model
configuration. Due to the padding, the number of weights and hence
memory savings are not directly proportional to the sparsity. Nonetheless,
DEEP R models with $\unit[1]{\%}$ connectivity still require $11\times$
fewer weights than if they were implemented using dense matrices and
those with $\unit[5]{\%}$ connectivity $4\times$ fewer.

Overall, in GeNN, connection sparsity significantly reduces training
times and memory usage. This represents a significant advantage compared
to SNN simulation libraries such as SNNtorch~\cite{eshraghian2021training},
NORSE~\cite{norse2021} or JAXsnn~\cite{muller_jaxsnn_2024} which
are built on standard Machine Learning toolkits like PyTorch or JAX
and thus only provide efficient support for dense weight matrices.

\begin{figure}[H]
\includegraphics{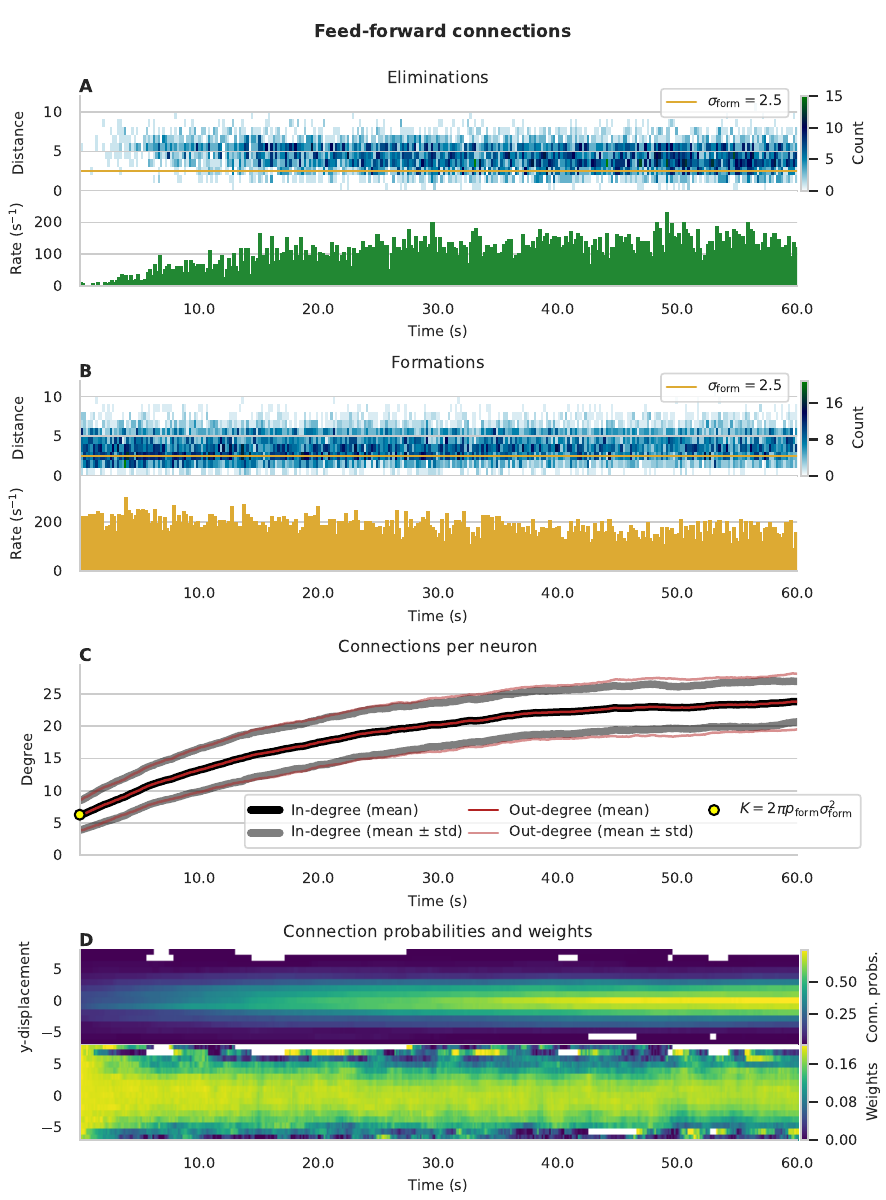}

\caption{\textbf{Evolution of feed-forward connections in topographic map model.
(A)}~Eliminations. Top: Color-coded count of removed connections
at given spatial separation in time bins of $\unit[200]{ms}$ and
spatial bins of $1$ unit. Standard deviation of the Gaussian-shaped
formation probability $\sigma_{\text{form}}$ indicated by yellow
line. Bottom: Histogram of elimination rate using time bins of $\unit[200]{ms}$.
\textbf{(B)}~Formations. Same display as in (A).\textbf{ (C)}~In-degree
(black) and out-degree (red); mean values (dark) and mean $\pm$ standard
deviation (light). Theoretical estimate for initial degree indicated
by circular marker. \textbf{(D)}~Connection probabilities and weights
for zero x-displacement. Initial and final frames correspond to \figref{topomap_methods}C.
\label{fig:topomap_connectivity_evolution_ff}}
\end{figure}

\begin{figure}[H]
\includegraphics{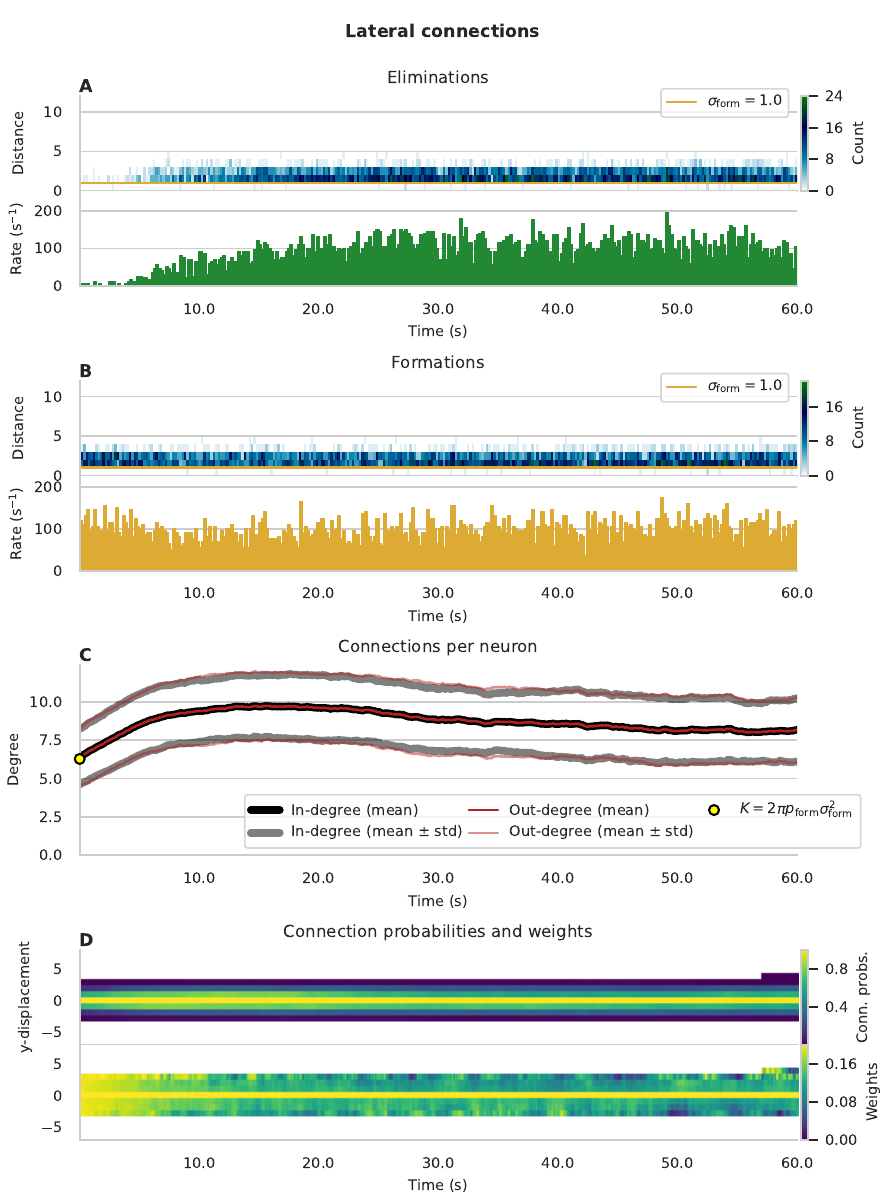}\caption{\textbf{Evolution of lateral connections in topographic map model.}
\textbf{(A)--(D)}~Same display as in \figref{topomap_connectivity_evolution_ff}.
Initial and final frames in panel (D) correspond to \figref{topomap_methods}D.
\label{fig:topomap_connectivity_evolution_lat}}
\end{figure}

\subsection{Topographic map formation}

Figures~\ref{fig:topomap_connectivity_evolution_ff} and \ref{fig:topomap_connectivity_evolution_lat}
show the evolution of feed-forward and lateral connections, respectively.
Following the rough topographic map initializion according to the
formation probability given in \tabref{topomap_model_description},
the elimination rate gradually increases and then stabilizes, while
the formation rate is mostly stable over the time course of the simulation,
see panels A and B. The increased elimination probabilities in our
implementation compared to~\cite{Bamford2010} result in similar
elimination and formation rates after the initial warm-up phase without
the need for fixing the maximum in-degree. These rates are higher
for the feed-forward than for the lateral connections due to the higher
standard deviation of the Gaussian-shaped formation probability. In
particular in the beginning of the simulation, it can be observed
that connections established across larger distances get predominantly
eliminated because the synaptic plasticity weakens these synapses
triggered by the spatially correlated Poisson input. Note that the
upper displays of panels A and B only count the changes in connectivity
at certain spatial separations of the respective grid locations without
taking into account the number of possible connections per distance.
For instance, there is only one grid position at zero distance (for
lateral connections this corresponds to autapses~\cite{Senk2022})
and four grid positions at a distance of one. Panel C shows how the
connections per neuron evolve on average. The results for in-degrees
and out-degrees are similar. The starting point matches the theoretical
expectation, and it is the same for feed-forward and lateral because
in both cases the parameters are chosen such that $p_{\text{form}}\sigma_{\text{form}}=1$.
From this starting point, the number of feed-forward connections per
neuron grows until saturation. In contrast, the number of lateral
connections decreases after an initial increase until reaching a stable
value. While \figref{topomap_methods}C and D show connection probabilities
and weights averaged over displacements for each grid position just
for the initial and final state, panel D of figures~\ref{fig:topomap_connectivity_evolution_ff}
and \ref{fig:topomap_connectivity_evolution_lat} illustrate the temporal
evolution across all y-displacements with zero x-displacement. The
gradual refinement of the receptive fields is visible mostly close
to the ideal location. In more distant places new synapses get formed,
but their weights decrease from their high initial value due to the
normal synaptic weight plasticity, and they get eventually eliminated
again.

\begin{figure}[H]
\includegraphics{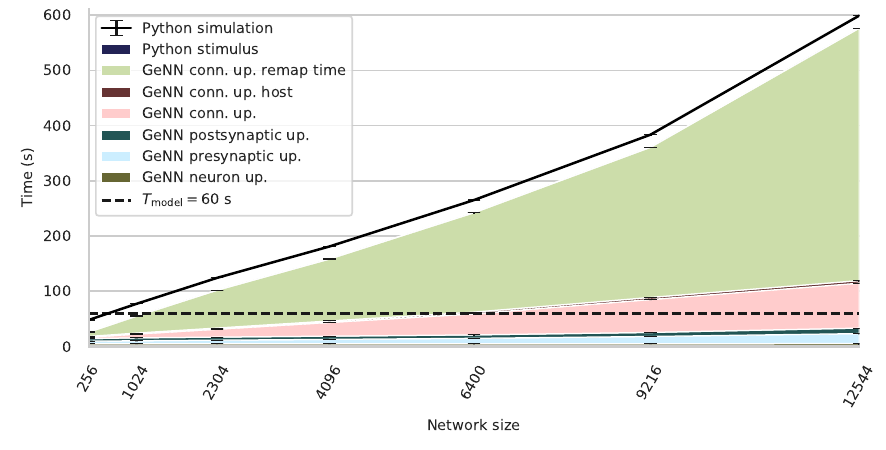}\caption{\textbf{Scaling performance of topographic map model. }Measured\textbf{
}wall-clock time of simulations with network sizes increasing from
scale $s=1$ ($16\times16=256$ neurons) to $s=7$ ($112\times112=12,544$
neurons); averages over three random seeds. Total simulation time
measured on the Python level (black curve) and stacked, color-coded
contributions resolved with GeNN built-in timers and a Python-level
timer for updating the stimulus center location of the correlated
input. Realtime indicated by black dashed line. \label{fig:topomap_performance}}
\end{figure}

We then measure the computational speed of the implementation. The
runtime for increasing network sizes is illustrated in \figref{topomap_performance}.
At the network size used in the original publication \cite{Bamford2010}
($s=1$) we observe faster-than-realtime performance with the GeNN
implementation. Even though the measured wall-clock time increases
when increasing the network size, the proportionality factor is smaller
than one: For example, at $s=7$ the network size grows by a factor
of $49$, but the simulation time only by ca. $10$. The largest
contributions to the total runtime are in fact due to the connectivity
updates, in particular the time for remapping the inverse ragged matrix,
followed by the update on the device required for STDP, while the
time for the host update is negligible. The number of successful rewiring
attempts remains constant when increasing the network size because
the formation rule only establishes local connections. In contrast,
the total number of rewiring attempts increases with the network size
but is performed in parallel. Neuron and synapse updates (i.e., presynaptic
and postsynaptic, including synaptic plasticity), and the Python code
to switch the stimulus center location of the correlated input are
also comparatively insignificant for larger network sizes. The gap
between the sum of individual contributions and the total simulation
time measured on the Python level is most likely due to the overheads
of Python calling C++.

\section{Discussion}

\label{sec:discussion}We propose a flexible and efficient structural
plasticity framework for the GeNN simulator which addresses the challenges
of parallelizing structural plasticity algorithms and handles their
complex dependencies on other neuronal and synaptic mechanisms, while
minimizing the computational cost of changing connectivity in allocated
data structures. We demonstrate this framework using two different
applications: In a spike-based ML context, we train recurrent classifiers
with very sparse connectivity using supervised learning and the DEEP
R structural plasticity rule. In a computational neuroscience context,
we show unsupervised learning leading to the refinement of a topographic
map using a structural plasticity rule with activity-dependent and
activity-independent contributions.

The `lottery ticket hypothesis'\cite{frankle_lottery_2019} states
that, within dense networks, there exist sparse sub-networks (the
winning tickets) which can be trained in isolation to obtain the same
test accuracy as the dense network in the same number of epochs. On
this basis, the sparse classifiers we trained using DEEP R appear
to be promising winning tickets as they maintain the same performance
as networks with dense connectivity, even though they are extremely
sparse: each hidden neuron only has around $5$ outgoing recurrent
synapses and each input only around $25$ feed-forward synapses. However,
as \figref{DVS-gesture-training}A shows, these potential winning
tickets converge more slowly than the dense networks. This reduced
convergence speed could be due to `competition' between e-prop learning,
L1 regularization and the rewiring process (which occurs every batch)
and it would be interesting to explore this further by retaining the
connectivity structure of the sparse networks trained using DEEP R,
reinitializing their weights and retraining them from scratch while
keeping the connectivity static. \textcite{evci_rigging_2020} found
that forming new connections randomly -- as DEEP R does -- limits
the performance of sparse networks and propose using gradients from
\emph{potential }synapses to guide where new synapses are formed.
While this approach is impossible in a truly sparse framework such
as we are developing here, it would be interesting to combine DEEP
R with more biologically-inspired `guidance' such as the distance-dependent
formation rule used in the topographic map formation model (\tabref{topomap_model_description}).
Additionally, exploring how different tasks affect the efficacy of
DEEP R and whether it can be combined with more efficient learning
rules such as Eventprop~\cite{wunderlich_event-based_2021,Nowotny2022}
would both be interesting future directions for this work.

Using the novel structural plasticity framework, we have implemented
a variation of the structural plasticity mechanism proposed by \textcite{Bamford2010}
which includes an interplay between structural plasticity, synaptic
plasticity and spatial inter-neuron distance dependence. The mechanism
promotes a refinement of receptive fields in a topographic map model~\cite{Bamford2010}
compatible with the results of~\cite{Bogdan2018}. Our analysis reveals
further insights into the time course of synapse formation and elimination,
in particular the feasibility of compensating for not fixing the maximum
in-degree by adjusting the elimination probabilities. While the SpiNNaker
implementation \cite{Bogdan2018} achieves realtime performance with
a simulation time-step of $h=\unit[1]{ms}$, GeNN demonstrates faster-than-realtime
performance despite using a $10\times$ smaller time step. In contrast
to previous implementations, the GeNN model applies multiple rewiring
attempts in parallel, and demonstrates the scaling performance on
a GPU upon upscaling the network size. At the original size, $10$
rewiring attempts are performed in parallel and at the largest tested
size $10\times49=490$. Other synaptic plasticity rules that involve
more inherent parallelism would benefit even more from this feature.
Both the structural plasticity mechanism and the network model used
to demonstrate topographic map refinement serve as proofs-of-concept
for the technological capability rather than claiming full biological
realism. \textcite{Bamford2010} and \textcite{Bogdan2018} provide
extensive discussions on the limitations and on potential improvements,
for example, axon branching could be guided by the existence of axons,
homeostatic mechanisms could avoid the need for new synapses to start
strong, and more realistic inputs, distance-dependent delays and inhibition
could be added. Any biological interpretation of our example results
should take the simplifications of the current approach into account.

The performance impact of structural plasticity updates on the two
models we simulate is drastically different. In our e-prop classifier
models, DEEP R only adds a $\unit[0.5]{\%}$ overhead to training
time whereas, when simulating large topographic map models (scale
$s=7$), structural plasticity updates account for around $\unit[90]{\%}$
of the simulation time. There are numerous reasons for this. Firstly,
DEEP R is only run after each batch ($18456$ timesteps for DVS-gesture)
whereas the refinement rule used in the topographic map model is run
much more frequently (every $10$ timesteps). Additionally, in each
of these batches, $512$ model instances are updated in parallel further
amortizing the cost of the DEEP R updates. The final issue is the
large `remap time' seen in \figref{topomap_performance} (see \subsecref{custom_conn_updates}).
Because, only around $\unit[50]{\%}$ of connectivity updates in the
topographic map model result in connection changes, one easy optimisation
would be to only perform the remapping process if the connectivity
has actually changed. Furthermore, more efficient GPU STDP implementations
have been recently developed~\cite{bautembach_even_2021} which do
not require this type of data structure at all. However, incorporating
further optimisations into GeNN is beyond the scope of this paper
and, while these overheads may limit the practicability of our approach
for large models using STDP with a postsynaptic spike-triggered term,
the performance impact of structural plasticity will be minimal for
many other models.

This study demonstrates the feasibility of expressing different structural
plasticity rules~\cite{Bamford2010,Bellec2018} using the syntax
of our novel structural plasticity framework and enabling efficient
GPU-based simulations of network dynamics. We hope that the proposed
framework serves as a useful tool to support further explorations
of biologically founded structural plasticity mechanisms and their
exploitation for spike-based ML. While we have not explored it here,
this framework could also be used to implement neurogenesis by instantiating
the final number of neurons at the start of the simulation without
any synaptic connections and connecting them over time. This approach
has been widely used to implement neurogenesis both in software simulators~\cite{huang_neurogenesis_2023}
and neuromorphic hardware~\cite{Hajizada2025}.

Despite considering generic challenges of structural plasticity implementations,
our design is tailored towards GPU-based systems and in particular
the GeNN simulator. However, such frameworks are potentially applicable
to other simulation software that uses code generation such as Brian
2~\cite{Stimberg2019} or NESTML~\cite{linssen_nestml_2025} as
well as programmable neuromorphic hardware. For example, ongoing work~\cite{schmidt_extension_2023}
to target the SpiNNaker~\cite{Furber2014} and SpiNNaker 2~\cite{gonzalez_spinnaker2_2024}
neuromorphic systems using NESTML would provide an interesting route
for running flexible structural plasticity rules at scale. Furthermore,
both Loihi~\cite{Davies2018} and BrainScaleS--2~\cite{pehle_brainscales-2_2022}
supports sparse connectivity and, potentially, code could be generated
to run on the general processors present in both systems to modify
this connectivity. Accelerated-time systems such as BrainScaleS--2
would be a particularly interesting choice for implementing structural
plasticity as they can overcome the long simulation times otherwise
required to simulate the slow speed of structural changes in biology.

\section*{Acknowledgements}

This project has received funding from EPSRC (grant numbers EP/V052241/1
to JK and EP/S030964/1 to TN and JK), the European Union’s Horizon
2020 Framework Programme for Research and Innovation under Specific
Grant Agreement No. 945539 (Human Brain Project SGA3) to TN and JS,
the European Union’s Horizon Europe Programme under the Specific Grant
Agreement No. 101147319 (EBRAINS 2.0 Project) to JS, and the VolkswagenStiftung
under the \textquotedbl STRUCTICITY: Multi-level STRUCTure and plasTICITY
in large-scale spiking neural networks for event-based sensing\textquotedbl{}
project (9D558/-1) to JS.

Compute time was provided through Gauss Centre for Supercomputing
application number 21018 and EPSRC (grant number EP/T022205/1) and
local GPU hardware was provided by an NVIDIA hardware grant award.

Johanna Senk is grateful to the Institute for Advanced Simulation
(IAS-6), Jülich Research Centre, for making a research visit at the
University of Sussex possible during which this project has been started,
and for providing access to their NVIDIA DGX A100 system used partially
for the research performed.

\printbibliography

\end{document}